\documentclass[runningheads]{llncs}
\usepackage[T1]{fontenc}
\usepackage{hyperref}
\usepackage{graphicx}
\usepackage{color}

\usepackage{amsfonts,amssymb}
\usepackage{bm}
\usepackage{booktabs}
\usepackage{cite}
\usepackage{multicol,multirow}
\usepackage{tabularx}
\usepackage{ulem}
\usepackage{orcidlink}

\newcommand{\orcid}[1]{{\orcidlink{#1}}}
\newcommand{\visited}{last accessed date: March 1, 2025}
\newcommand{\shortname}{FoodMLLM-JP}
\newcommand{\revised}[1]{\textcolor{blue}{#1}}
\definecolor{orcidlogocol}{HTML}{A6CE39}

\newcommand{\tablefontsize}{\scriptsize}
\newcolumntype{t}{>{\tablefontsize}l}
\newcolumntype{s}{>{\tablefontsize}c}
\newcolumntype{d}{>{\tablefontsize}r}
\newcolumntype{L}{>{\raggedright\arraybackslash}X}
\newcolumntype{C}{>{\centering\arraybackslash}X}
\newcolumntype{R}{>{\raggedleft\arraybackslash}X}
\newcolumntype{T}{>{\tablefontsize}L}
\newcolumntype{S}{>{\tablefontsize}C}
\newcolumntype{D}{>{\tablefontsize}R}

\begin{document}
\title{\shortname: Leveraging Multimodal Large Language Models for Japanese Recipe Generation}

\titlerunning{\shortname}

\author{Yuki Imajuku\,\orcid{0009-0005-7470-8305}\inst{1} \and Yoko Yamakata\,\orcid{0000-0003-2752-6179}\inst{1} \and Kiyoharu Aizawa\,\orcid{0000-0003-2146-6275}\inst{1}}

\authorrunning{Y. Imajuku et al.}

\institute{The University of Tokyo, Tokyo, Japan\\\email{\{imajuku,yamakata,aizawa\}@hal.t.u-tokyo.ac.jp}}

\maketitle  

\begin{abstract}
Research on food image understanding using recipe data has been a long-standing focus due to the diversity and complexity of the data.
Moreover, food is inextricably linked to people's lives, making it a vital research area for practical applications such as dietary management.
Recent advancements in Multimodal Large Language Models (MLLMs) have demonstrated remarkable capabilities, not only in their vast knowledge but also in their ability to handle languages naturally.
While English is predominantly used, they can also support multiple languages including Japanese.
This suggests that MLLMs are expected to significantly improve performance in food image understanding tasks.
We fine-tuned open MLLMs LLaVA-1.5 and Phi-3 Vision on a Japanese recipe dataset and benchmarked their performance against the closed model GPT-4o.
We then evaluated the content of generated recipes, including ingredients and cooking procedures, using 5,000 evaluation samples that comprehensively cover Japanese food culture.
Our evaluation demonstrates that the open models trained on recipe data outperform GPT-4o, the current state-of-the-art model, in ingredient generation.
Our model achieved F1 score of 0.531, surpassing GPT-4o's F1 score of 0.481, indicating a higher level of accuracy.
Furthermore, our model exhibited comparable performance to GPT-4o in generating cooking procedure text.

\textit{(We found errors in the calculation of evaluation metrics, which were corrected in this version with \revised{modifications highlighted in blue}.
Please also see the Appendix.)}

\end{abstract}

\section{Introduction}\label{sec:introduction}
The task of understanding food images such as estimating dish names and ingredients from food images has been an active area of research, particularly within the context of leveraging recipe data~\cite{recipe1m,recipe1mp,inverse_cooking,recipe_program,l_wang2023,fire,car,foodlmm}.
The ability to extract information from food images has promising applications in personalized dietary management, such as nutrient estimation and the identification of potential allergens.
Given the profound connection between dietary habits and individual well-being, research in this domain holds substantial importance.

The realm of image understanding and captioning has witnessed remarkable progress in recent years, driven by the advent of Multimodal Large Language Models (MLLMs)~\cite{gpt-4,gemini-15,claude3}.
While these powerful models are accessible via APIs, they remain closed, incurring usage fees and obscuring their underlying technical details.
Training Large Language Models (LLMs) typically demands vast computational resources, rendering individual training efforts challenging.
However, the release of open LLMs, represented by Meta's Llama~\cite{llama}, has democratized access to these models~\cite{llama2,llama3,phi3_vision,gemma,mistral,qwen,yi}.
This has led to extensive research on building MLLMs utilizing these models, leading to the availability of locally deployable MLLMs~\cite{phi3_vision,pali_gemma,llava,llava_16,qwen_vl,cogvlm,share_gpt4v}.
Consequently, research on leveraging these open MLLMs for specific domains is gaining momentum~\cite{llava_med,geo_chat,foodlmm}.

\begin{figure}[t]
\centering
\includegraphics[width=0.71\textwidth]{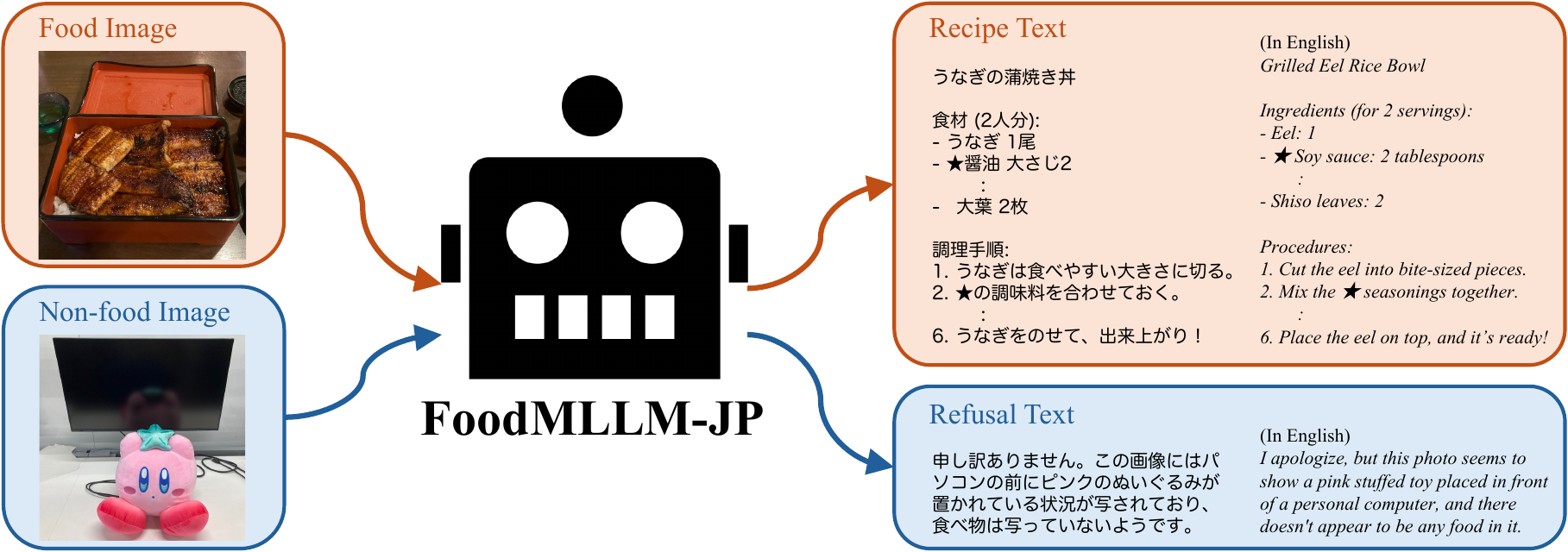}
\caption{Overview of our models. \textbf{\color[rgb]{0.753,0.310,0.082} Up}: the example of generated recipe text from input food image. \textbf{\color[rgb]{0.133,0.369,0.600} Down}: the example of generated refusal text from input non-food image. Both of them are in Japanese and generated from a our model.}\label{fig:overview}
\end{figure}

In this research, we focus on the task of generating recipe text from food images.
We conduct a comprehensive investigation into the capabilities of MLLMs in understanding food by performing extensive evaluations of the recipes generated from food images, comparing different methods and training data for fine-tuning MLLMs, and providing a holistic analysis from MLLM training to evaluation.
Furthermore, this research marks the first exploration of recipe generation tasks in a non-English language (Japanese) by utilizing the Rakuten Recipe dataset~\cite{rakuten_recipe}, a Japanese recipe dataset.
To evaluate MLLMs from the perspective of understanding Japanese food culture, we aim to assess a diverse range of meals equitably.
We created a new 50-category evaluation scheme based on meal types (e.g., staple food, main dishes) and main ingredients (e.g., meat, fish), using 5,000 recipes.
Moreover, unlike previous recipe generation research, we preserve the original text without performing normalization processing on ingredients or cooking procedure descriptions. In conjunction with this, we propose a novel evaluation methodology to evaluate free-form ingredient lists using LLMs.
Additionally, taking advantage of the versatility of MLLMs, we explore a new approach by incorporating non-food images and their captions during training.
This allows the model to determine whether an input image is a food image before generating recipe text.
Figure \ref{fig:overview} describes this feature.
This approach is significant as it allows the model to handle undesirable or malicious inputs in real-world applications without requiring a separate model.
Experiments show that our fine-tuned open MLLMs on recipe data achieve an F1 score of 0.531 for ingredient lists, outperforming the closed MLLM GPT-4o with an F1 score of 0.481 in accurately estimating used ingredients.
In terms of cooking procedure text generation, we achieves sacreBLEU score of \revised{13.69}, comparable to GPT-4o's score of \revised{8.22}.
Our contributions can be summarized in following three points:
\begin{description}
\item[Comprehensive Pipeline] We present a comprehensive pipeline that includes the preparation for fine-tuning open-source MLLMs to evaluation based on curated evaluation data that considers food culture. 
Additionally, we conduct the first attempt to evaluate the recipe generation capability from food images in a non-English language (Japanese).
\item[Diverse Data] Leveraging the versatility of MLLMs, we retain the original recipe text as created by humans, while also incorporating non-food images and their captions into the training process.
This approach introduces greater data diversity compared to previous recipe generation studies.
We observed that, for certain MLLMs, increasing the data even with non-recipe content can lead to performance improvements.
\item[Fine-tuning Insights] Through the task of recipe text generation from food images, we analyze the performance differences caused by different base MLLMs and adjusting parameters of MLLMs during fine-tuning.
We demonstrate that, with specific fine-tuning methods, it's possible to achieve performance surpassing that of a high-performing closed MLLM GPT-4o.
\end{description}

\section{Related Work}\label{sec:related_work}
\subsection{Food Computing with Recipe Data}
The Recipe1M dataset~\cite{recipe1m}, containing approximately 1M recipes, has been extensively utilized in research exploring deep learning techniques for food image understanding.
Notable works include Marin et al.~\cite{recipe1mp}, that proposed a cross-modal retrieval method for food images and recipe text, Salvador et al.~\cite{inverse_cooking}, that demonstrated a recipe generation pipeline from food images by first estimating ingredients and then generating cooking instructions and showed the superiority of generated recipes in both quantitatively and qualitatively, Papadopoulos et al.~\cite{recipe_program}, that embedded recipe text and food images into a shared feature space to generate pseudo-programs representing cooking instructions, and Chhikara et al.~\cite{fire}, that improved ingredient and cooking instruction generation by utilizing generated recipe titles and ingredient lists as input to a language model.
These studies utilize data that has undergone normalization processes for ingredients and cooking procedures, as proposed in Inverse Cooking~\cite{inverse_cooking}.
For example, similar ingredients like ``gorgonzola cheese'' or ``cheese blend'' are grouped into ``cheese,'' a single ingredient category.
This practice, while simplifying data handling, diminishes the diversity of expression in the data.

These days, research utilizing LLMs and MLLMs in food computing has also been emerging.
Salvador et al. enhanced recipe retrieval performance by expanding the data to include two additional sources: image segments using SAM~\cite{sam} and a LLM-generated visual description imagined from the recipe text~\cite{car}.
Yin et al. developed a MLLM-based conversational assistant with a dataset encompassing multiple food-related tasks, including recipe generation~\cite{foodlmm}.

Shifting our focus to Japanese recipe datasets, there are Rakuten Recipe Dataset~\cite{rakuten_recipe}, which provides about 800K recipes from Rakuten Recipe\footnote{\url{https://recipe.rakuten.co.jp/}, \visited}, and Cookpad Dataset~\cite{cookpad}, which offers approximately 1.72M recipes from Cookpad\footnote{\url{https://cookpad.com/}, \visited}.
These datasets are valuable from the perspective that they utilize photos of food prepared in everyday households, making them more representative of those used in practical applications such as dietary management.
Despite such rich datasets, only a few studies have been conducted on multimodal exploitation~\cite{l_wang2023}.

\subsection{MLLMs}
Prevalent LLMs~\cite{llama,llama2,llama3,phi3_vision,gemma,mistral,qwen,yi} are causal language models, which process input sequences $\bm{x}_{t}$ by tokenizing them according to a predefined vocabulary and predicting the distribution of probabilities for the next token's occurrence.
Despite minor variations, most of these models employ Transformer architecture~\cite{transformer}, which embeds each token $t_{i}$ into a $d$-dimensional feature vector $\bm{z}_{i} \in \mathbb{R}^{d}$ before feeding it into Transformer layers.

Since LLMs are trained solely on language data, they cannot directly process non-language information such as images or audio.
To enable them to handle multimodal inputs, various studies attempted to extend LLMs~\cite{phi3_vision,pali_gemma,llava,llava_16,qwen_vl,cogvlm,share_gpt4v}.
A common approach involves extracting features $\bm{h}_{m}$ from non-language modal information $\bm{x}_{m}$ using a feature extractor $\mathcal{E}_{m}$, transforming them into suitable features for LLMs input via a mapping function $f$, and feeding them into Transformer layers of LLMs alongside the text token embeddings $\bm{z}_{i}$.
Particularly for image modal extension, image encoders are predominantly based on Vision and Language Models (VLMs) like CLIP~\cite{clip}.
While minor variations exist in training data, image feature sequence input methods, and other aspects, the prevailing approach involves training the mapping function $f$ on a mixed dataset of images and text in the first step, followed by instruction tuning in the next step.
LLaVA~\cite{llava,llava_16} is one of the well-known open MLLMs, and Phi-3 Vision~\cite{phi3_vision} has good features as a lightweight model, which is used in this research.

\section{\shortname}\label{sec:methods}
\subsection{Data Preparation}
\subsubsection{Recipe Data}
We utilized the Rakuten Recipe dataset~\cite{rakuten_recipe} for recipe data, which contains 796,274 recipes.
In addition to basic components like titles, ingredients, cooking instructions, and completed dish images, it includes information such as three-level categories that classify dishes.
We performed the following three operations on this dataset to construct 635,873 training data and 5,000 test data:\\

\noindent\textit{Step 1. Dataset Splitting} First, we divided the entire dataset by the top-level category and split the dataset within each category so that the ratio of training data to evaluation data was 4:1.
As a result, we obtained 638,997 training data and 157,277 test data.

\noindent\textit{Step 2. Exclusion of Recipes with Broken Image Files} We excluded a part of the dataset where the image could not be properly read, resulting in 635,873 training data and 156,522 test data.

\begin{figure}[t]
\centering
\includegraphics[width=0.8\textwidth]{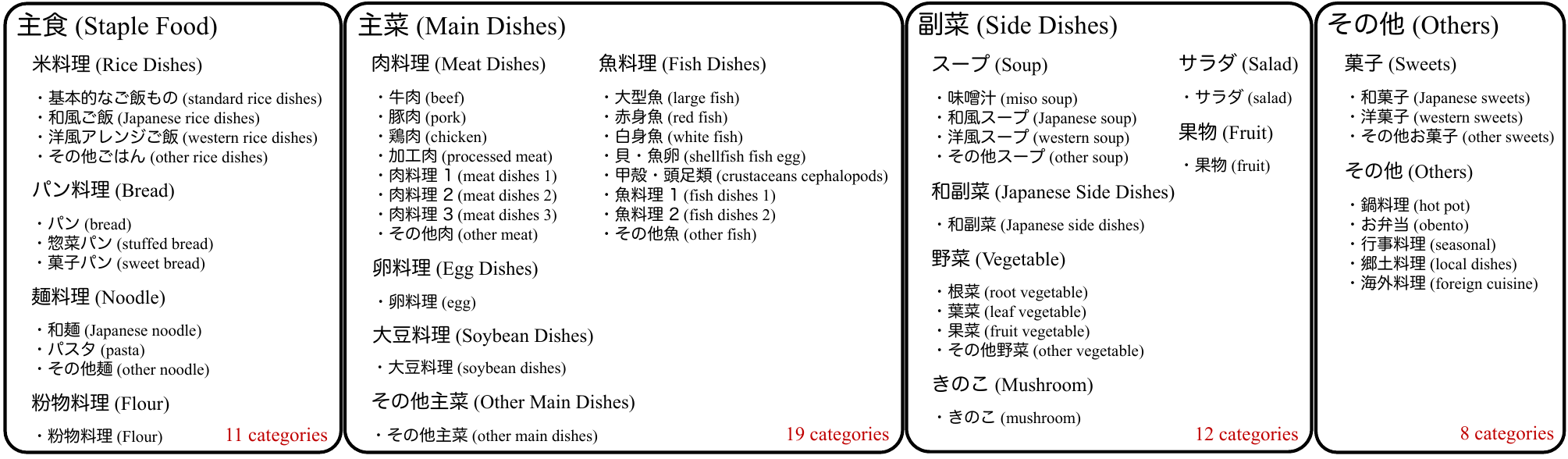}
\caption{50-categories we created for test data.}\label{fig:category}
\end{figure}

\noindent\textit{Step 3. Test Data Selection} Due to the uneven number of items between categories (e.g., many salad posts), we created 50 new categories that cover common foods in Japan by focusing on meal types as well as the main ingredients used.
Figure \ref{fig:category} lists all 50 categories.
The assignment from original top two-level categories in the dataset to the new categories was done manually.
Then, to reduce evaluation costs, we randomly sampled 100 test data from each category, resulting in a total of 5,000 test data.

\subsubsection{Non-Food Data}
We utilized the STAIR Captions dataset~\cite{stair} for captioning data of non-food images.
This dataset consists of 5 Japanese captions per image from the MS-COCO~\cite{mscoco} dataset.
To ensure the use of data other than food images, we excluded images containing objects with supercategories of ``kitchen'' and ``food''.
As a result, we extracted 63,223 images from the train set.  

\subsection{Recipe Generation Training}\label{subsec:train_data}
\begin{figure}[t]
\centering
\includegraphics[width=0.9\textwidth]{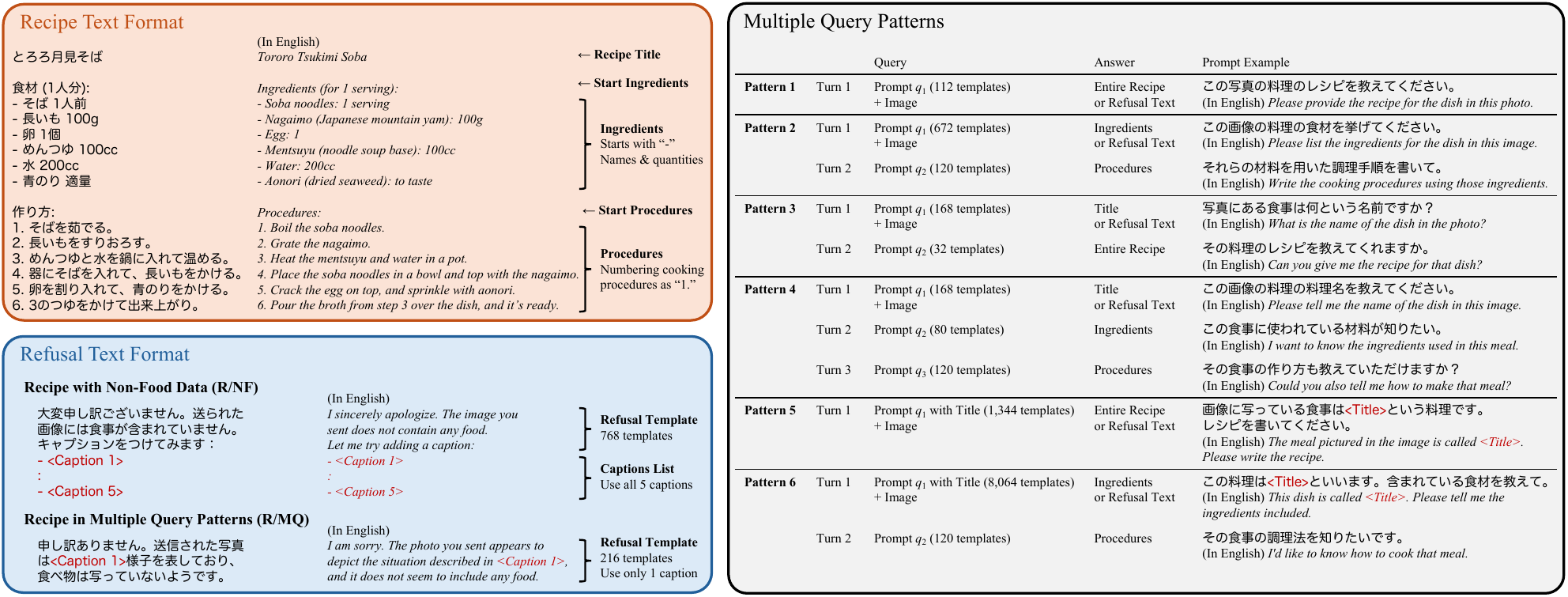}
\caption{Train data desciption. \textbf{\color[rgb]{0.753,0.310,0.082} Upper Left}: the format of recipes. \textbf{\color[rgb]{0.133,0.369,0.600} Lower Left}: the format of refusal text \textbf{Right}: description of six patterns and example user query prompts.}\label{fig:query}
\end{figure}

Recipe data typically includes text data such as the title, the ingredients used, and the cooking instructions, often accompanied by an image of the completed dish.
The format of the recipe text is shown in the upper left part of Figure \ref{fig:query}.
In this research, we fine-tune MLLMs using this data to enable the inference of the dish name, ingredients used, and cooking procedures from an input food image.
The model takes a template containing the completed image and a query text $q$ as a prompt from the user and learns desirable answer including the recipe text portion like SFT~\cite{instruct_gpt} and LLaVA~\cite{llava}.
This approach has the advantage of being practical due to the simple loss design by cross-entropy loss.
At the same time, since LLMs have vast knowledge and can generate natural Japanese text, it may enables more diverse and accurate recipe generation.
In this reseearch, we compare and examine the following three ways for recipe learning:

\begin{description}
\item[Recipe (R)] Only recipe data is used for training models.
The input is the completed image with $q$ = ``'' (empty), and the output is recipe text.
The model is trained on recipe data only, and the output is the recipe text.
\item[Recipe with Non-Food Data (R/NF)] Non-food data is used for training models in addition to the recipe data.
The input is the image with $q$ = ``'' (empty), the same as above.
The output is the recipe text for food images, and an apology message with the image caption for non-food images.
The format of the refusal text is shown in the lower left part of Figure \ref{fig:query}.
\item[Recipe in Multiple Query Patterns (R/MQ)] The model is trained on the same data as R/NF, but with multiple query $q$ and answer patterns.
There are six query patterns, including requesting the entire recipe, title only, ingredients only, procedures only, or patterns where the recipe title is given along with the food image.
Examples of these patterns are shown on the right side of Figure \ref{fig:query}.
As shown in the lower left of Fig. \ref{fig:query}, the format of the response refusing to generate a recipe is slightly different from that for R/NF.
Instead of using all five captions in the dataset, one of them is used in the apology message.
This is because the five captions are similar, and the trained model often repeatedly outputs the same caption for five times.
\end{description}

\subsection{Evaluation with LLMs}
Accurately evaluating the content of generated recipes is challenging due to inherent ambiguities, such as subtle differences in ingredient types or variations in the order of listing.
To address this, Salvador et al.~\cite{inverse_cooking} performed ingredient normalization and removed infrequent ingredients during training.
However, this method sacrifices recipe diversity and deviates from real-world scenarios.
Furthermore, evaluating Japanese recipes presents additional difficulties due to variations in ingredient representation, such as kanji and hiragana, and different expressions for the same ingredient, like ``gohan'' (rice) and ``hakumai'' (white rice).
Nevertheless, recent LLMs are expected to be capable of considering these factors.
In fact, research has emerged that utilizes highest performance LLM, such as GPT-4, for automatic evaluation, circumventing the need for expensive human evaluation while assessing model performance~\cite{vicuna,alpaca_farm}.
In this research, we evaluate the performance of our model in an open setting, allowing it to generate recipes without pre-specifying ingredient or procedure classes.
This is achieved by having GPT-4o determine common and different ingredients between the generated recipe and the ground truth recipe.
We provide the GPT-4o model with two sets of ingredients: a generated ingredient set $\mathcal{S}_{1}$, and a ground-truth ingredient set $\mathcal{S}_{2}$.
Both sets are constructed from elements $s_i$ within the universal set of all possible text $\mathcal{T}$.
The model is then tasked with producing the intersection $\mathcal{S}_{1} \cap \mathcal{S}_{2}$ and the set differences $\mathcal{S}_{1} \setminus \mathcal{S}_{2}, \mathcal{S}_{2} \setminus \mathcal{S}_{1}$ of these two sets.
Moreover, we use GPT-4o to separately judge seasonings, which are difficult to estimate from the appearance of a dish, and other ingredients.
This allows for a more detailed analysis that was previously expensive and impractical to perform manually.

\begin{figure}[t]
\centering
\includegraphics[width=0.9\textwidth]{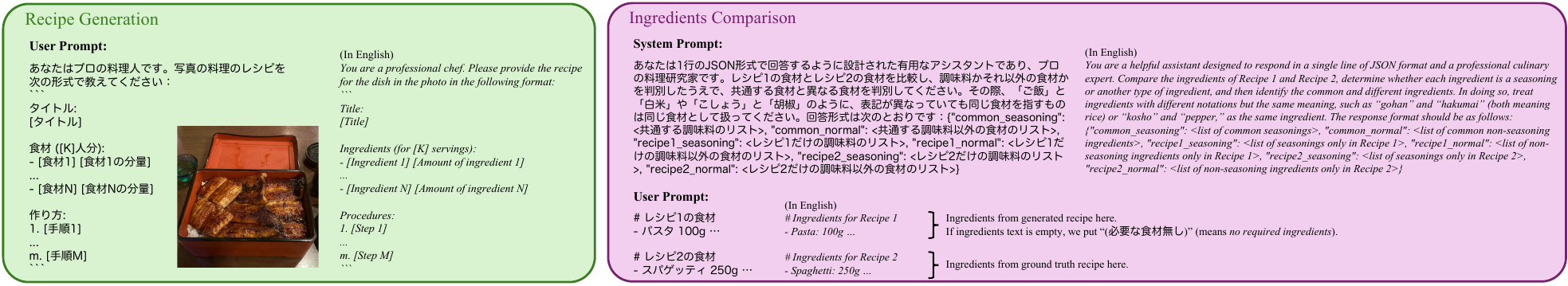}
\caption{The actual prompts used for GPT-4o inferences. \textbf{\color[rgb]{0.231,0.494,0.137} Left}: the prompt for recipe generation. \textbf{\color[rgb]{0.451,0.125,0.416} Right}: the prompt for ingredients comparison between generated recipe and ground truth. Both of them are in Japanese.}\label{fig:prompt}
\end{figure}

\section{Experiments}\label{sec:experiments}
\subsection{Closed Model Experiment Details}
We employed the OpenAI's GPT-4o model (\texttt{gpt-4o-2024-05-13}) as the closed MLLM.
We used the model to generate recipe from food images and evaluate ingredients via the OpenAI API.
The actual prompts used for each case are shown in Figure \ref{fig:prompt}.
In all cases, the sampling temperature was set to 0.0.

Firstly, We generate recipe texts using the 5,000 evaluation data prepared in Section \ref{sec:methods}.
The model was provided with a completed image of the recipe to be evaluated and a text instructing it to generate a recipe text in a specified format based on the image.
The image input size option was set to \texttt{"auto"}.

Secondly, we automatically compared the ingredients listed in the recipe text generated from the input image with the correct recipe using the GPT-4o model.
The ingredient list output by the model, the ingredient list of the correct recipe, and the instruction text were combined into a single string and input to the model.
The instruction was to identify common and different ingredients, determine whether they are seasonings or other ingredients, and output the results in JSON format.
By mechanically processing the response, we implemented calculations such as IoU and F1 score.

\subsection{Open Model Experiment Details}
We fine-tuned available open MLLMs using the three types of training data described in Section \ref{subsec:train_data}.
We selected the 7B and 13B models of LLaVA-1.5~\cite{llava_16} as base models, and the Phi-3 Vision~\cite{phi3_vision} as a more lightweight base model.
For the LLaVA-1.5 models, we fine-tuned the adapter $f$, which converts image features to input features for the LLM, with a learning rate of $2 \times 10^{-5}$ , and the LLM part with a learning rate of $2 \times 10^{-4}$ using LoRA~\cite{lora} $(r = 128)$ modules.
This configuration follows the hyperparameter settings recommended by the LLaVA authors.
For the Phi-3 Vision model, we take advantage of its lightweight nature to not only perform LoRA fine-tuning of the LLM part, but also experiment with other training methods: full parameter tuning of the LLM part, and fine-tuning the entire including CLIP vision encoder.
Note that the vision encoder is generally used with fixed weights, but we also fine-tune it, too.
In all cases, we used AdamW~\cite{adamw} as the optimization algorithm and set the batch size to 128 and the number of epochs to 1.
We also used a learning rate scheduler that linearly increases the learning rate for the first 3\% of training steps and then cosine decays.
We used 4x A100 80GB GPUs for training.
We summarize these settings in Table \ref{table:model_type}.
We also include the model names used in this paper for concise presentation of the experimental results.
During inference, we performed greedy sampling with a temperature parameter of 0.0 and generate tokens up to 2048 tokens.
We used a A100 40GB GPU for inference and \texttt{fp16} type.

\begin{table}[t]
\caption{Description of models in this paper and training hyperparameters. The word \texttt{lora}, \texttt{ft}, and \texttt{allft} in the model name indicates that the fine-tuning method is LoRA or full-parameter, and full-parameter includes the vision encoder, respectively. Also, the suffixes \texttt{nf} and \texttt{mq} after that indicate the type of training data.}\label{table:model_type}

\begin{tabularx}{\textwidth}{@{}tTTTTTt@{}} \toprule
 & & \multicolumn{2}{s}{LLM fine-tuning} & \multicolumn{2}{s}{Vision module} & \multicolumn{1}{s}{Training} \\\cmidrule(lr){3-4}\cmidrule(lr){5-6}
Model name & Base model & method & lr & Adapter lr & Encoder lr & \multicolumn{1}{s}{Data} \\\midrule
\texttt{gpt-4o} & GPT-4o & \multicolumn{5}{s}{(Not fine-tuned because of closed model)} \\\midrule
\texttt{llava7b-lora} & \multirow{3}{*}{\shortstack[l]{LLaVA-1.5\\(7B)}} & \multirow{3}{*}{\shortstack[l]{LoRA\\($r=128$)}} & \multirow{3}{*}{$2 \times 10^{-4}$} & \multirow{3}{*}{$2 \times 10^{-5}$} & \multirow{3}{*}{Freeze} & R \\
\texttt{llava7b-lora-nf} & & & & & & R/NF \\
\texttt{llava7b-lora-mq} & & & & & & R/MQ \\\midrule
\texttt{llava13b-lora} & \multirow{3}{*}{\shortstack[l]{LLaVA-1.5\\(13B)}} & \multirow{3}{*}{\shortstack[l]{LoRA\\($r=128$)}} & \multirow{3}{*}{$2 \times 10^{-4}$} & \multirow{3}{*}{$2 \times 10^{-5}$} & \multirow{3}{*}{Freeze} & R \\
\texttt{llava13b-lora-nf} & & & & & & R/NF \\
\texttt{llava13b-lora-mq} & & & & & & R/MQ \\\midrule
\texttt{phi3v-lora} & \multirow{3}{*}{Phi-3 Vision} & \multirow{3}{*}{\shortstack[l]{LoRA\\($r=128$)}} & \multirow{3}{*}{$2 \times 10^{-4}$} & \multirow{3}{*}{$2 \times 10^{-5}$} & \multirow{3}{*}{Freeze} & R  \\
\texttt{phi3v-lora-nf} & & & & & & R/NF\\
\texttt{phi3v-lora-mq} & & & & & & R/MQ \\\midrule
\texttt{phi3v-ft} & \multirow{3}{*}{Phi-3 Vision} & \multirow{3}{*}{\shortstack[l]{Full\\parameter}} & \multirow{3}{*}{$2 \times 10^{-4}$} & \multirow{3}{*}{$2 \times 10^{-5}$} & \multirow{3}{*}{Freeze} & R \\
\texttt{phi3v-ft-nf} & & & & & & R/NF \\
\texttt{phi3v-ft-mq} & & & & & & R/MQ \\\midrule
\texttt{phi3v-allft} & \multirow{3}{*}{Phi-3 Vision} & \multirow{3}{*}{\shortstack[l]{Full\\parameter}} & \multirow{3}{*}{$2 \times 10^{-4}$} & \multirow{3}{*}{$2 \times 10^{-5}$} & \multirow{3}{*}{$2 \times 10^{-5}$} & R \\
\texttt{phi3v-allft-nf} & & & & & & R/NF \\
\texttt{phi3v-allft-mq} & & & & & & R/MQ \\\bottomrule
\end{tabularx}
\end{table}

\subsection{Evaluation Metrics}
For evaluation, we used the 5,000 data samples carefully selected as described in Section \ref{sec:methods}.
First, to assess the model's training performance, we calculated the Perplexity~\cite{perplexity} against the ground truth recipes.
Second, to evaluate the content of the generated recipes, we examined the recipe format by counting the cases where the model refused to generate a recipe.
As LLMs sometimes exhibit repetitive generation, we evaluated how much of the recipe components (title, ingredients, and instructions) were correctly output when the model fell into an infinite loop.
Third, We treated incorrectly generated elements as empty strings of length 0, and then performed sacreBLEU~\cite{sacre_bleu} and ROUGE-L~\cite{rouge} evaluations for the cooking procedures.
We also evaluated ingredient content using GPT-4o.
For sacreBLEU and ROUGE-L calculations, we tokenized the procedures using Mecab~\cite{mecab} morphological analysis with IPAdic\footnote{Used this library: \url{https://pypi.org/project/ipadic/}, \visited} as the dictionary.

We first discuss the validity of using GPT-4o for ingredient judgment.  
We conducted a manual check on 100 data samples, balanced across 50 categories, from the 5,000 evaluated data samples for the ingredients generated by GPT-4o.
Out of 1,193 ingredients, 1,122 (94\%) were perfectly answered.
Of 71 incorrect answers, 45 were misjudgments of whether an ingredient was a seasoning or not. Specifically, most cases involved counting non-flavoring ingredients like salad oil, water, and flour as seasonings. However, some of these cases were difficult even for humans to interpret.
The remaining 26 cases were due to the inability to distinguish between ingredient expressions.
Specifically, there were many cases where kanji and hiragana expressions or different words expressing the same ingredients were not recognized as the same.
However, there were no ingredients that were completely wrong, and the accuracy rate of judging ingredients reached 98\%, leading us to conclude that this GPT-4o based metric is valid.

\begin{table}[t]
\caption{Table that summarizes the Perplexity calculated for ground truth recipes and the count of recipes in the correct format or not.}\label{table:result_format}

\begin{tabularx}{\textwidth}{@{}tSSSSss@{}} \toprule
 & & \multicolumn{5}{s}{Recipe format} \\\cmidrule(lr){3-7}
Model name & Perplexity & Completed & Refusal & Error title & Error ingredients & Error procedures \\\midrule
\texttt{gpt-4o} & --- & 5000 & 0 & 0 & 0 & 0 \\\midrule
\texttt{llava7b-lora} & 1.924 & 4930 & 0 & 2 & 21 & 47 \\
\texttt{llava7b-lora-nf} & 1.876 & 4940 & 1 & 0 & 18 & 41 \\
\texttt{llava7b-lora-mq} & 1.962 & 4951 & 0 & 0 & 14 & 35 \\
\texttt{llava13b-lora} & 1.895 & 4940 & 0 & 1 & 26 & 33 \\
\texttt{llava13b-lora-nf} & 1.861 & 4927 & 0 & 2 & 30 & 41 \\
\texttt{llava13b-lora-mq} & 1.971 & 4945 & 1 & 0 & 21 & 33 \\\midrule
\texttt{phi3v-lora} & 1.861 & 4970 & 0 & 0 & 13 & 17 \\
\texttt{phi3v-lora-nf} & 1.858 & 4968 & 0 & 0 & 15 & 17 \\
\texttt{phi3v-lora-mq} & 1.970 & 4957 & 0 & 3 & 21 & 19 \\
\texttt{phi3v-ft} & 1.735 & 4964 & 0 & 0 & 19 & 17 \\
\texttt{phi3v-ft-nf} & 1.740 & 4983 & 1 & 0 & 7 & 9 \\
\texttt{phi3v-ft-mq} & 1.904 & 4964 & 1 & 0 & 17 & 18 \\
\texttt{phi3v-allft} & 1.731 & 4975 & 0 & 1 & 12 & 12 \\
\texttt{phi3v-allft-nf} & 1.731 & 4962 & 2 & 0 & 16 & 20 \\
\texttt{phi3v-allft-mq} & 1.876 & 4968 & 3 & 0 & 9 & 20 \\\bottomrule
\end{tabularx}
\end{table}

\subsection{Results}
We present the evaluation results of the generated recipes.
First, Table \ref{table:result_format} shows Perplexity calculated against the ground truth 5,000 recipes and the statistics of how accurately the recipes were output in the correct format.
We could not calculate Perplexity for GPT-4o because it is a closed model.
The results show that GPT-4o can generate recipes that perfectly match the format, even though it is 0-shot and only specified by text instructions.
It also never falls into a loop, suggesting its high language capabilities.
Next, we focus on the results of fine-tuning the open models.
Looking at Perplexity, we find that Phi-3 Vision, trained the entire model including the image encoder, performs the best, while LLaVA-1.5 7B LoRA fine-tuned models perform the worst.
However, it is important to note that this metric does not directly indicate the quality of recipe generation.
Focusing on the differences in the training data, we observe that models trained on R/MQ data tend to have the worst Perplexity overall.
Also, while the Perplexity of the LLaVA-1.5 model improves when trained with additional non-food data, the Phi-3 Vision model shows almost no change.
Looking at the format of the generated recipes, we see that errors occur in about 1\% of cases.
However, it is clear that most of the errors are due to failures in generating ingredients or cooking instructions, rather than mistakenly recognizing the image as non-food.

\begin{table}[t]
\caption{Evaluation of ingredients and procedure texts comparison between models. The underlined number indicates the best score and the dotted underlined number indicates the second-best score. For ingredients, the scores are presented as overall score (non-seasoning score / seasoning score).}\label{table:result_ingproc}

\begin{tabularx}{\textwidth}{@{}tSSSdd@{}} \toprule
 & \multicolumn{3}{s}{ingredient (evaluated by GPT-4o)} & \multicolumn{2}{s}{procedure}\\\cmidrule(lr){2-4}\cmidrule{5-6}
Model name & micro F1 & micro Precision & micro Recall & \tiny sacre\scriptsize BLEU & \tiny ROUGE-\scriptsize L \\\midrule
\texttt{gpt-4o} & 0.481(0.470/0.495) & 0.451(0.442/0.463) & \uline{0.515}(0.501/\uline{0.532}) & \revised{8.22} & \revised{41.72} \\\midrule
\texttt{llava7b-lora} & 0.470(0.472/0.467) & 0.498(0.516/0.479) & 0.444(0.434/0.456) & \revised{8.83} & \revised{43.98} \\
\texttt{llava7b-lora-nf} & 0.478(0.483/0.472) & 0.501(0.521/0.480) & 0.457(0.450/0.464) & \revised{9.41} & \revised{44.46} \\
\texttt{llava7b-lora-mq} & 0.486(0.496/0.475) & 0.507(0.532/0.481) & 0.466(0.464/0.469) & \revised{10.04} & \revised{45.02} \\
\texttt{llava13b-lora} & 0.476(0.481/0.470) & 0.502(0.520/0.481) & 0.453(0.448/0.460) & \revised{9.37} & \revised{44.61} \\
\texttt{llava13b-lora-nf} & 0.488(0.500/0.472) & 0.514(0.540/0.484) & 0.464(0.466/0.461) & \revised{9.99} & \revised{44.86} \\
\texttt{llava13b-lora-mq} & 0.484(0.492/0.474) & 0.505(0.527/0.480) & 0.464(0.461/0.469) & \revised{10.14} & \revised{45.00} \\\midrule
\texttt{phi3v-lora} & 0.447(0.440/0.456) & 0.476(0.482/0.468) & 0.422(0.405/0.444) & \revised{10.08} & \revised{45.01} \\
\texttt{phi3v-lora-nf} & 0.447(0.442/0.454) & 0.472(0.480/0.462) & 0.425(0.409/0.446) & \revised{9.98} & \revised{44.86} \\
\texttt{phi3v-lora-mq} & 0.438(0.431/0.447) & 0.465(0.472/0.457) & 0.415(0.396/0.438) & \revised{9.63} & \revised{44.49} \\
\texttt{phi3v-ft} & 0.495(0.500/0.490) & 0.518(0.537/0.498) & 0.474(0.468/0.481) & \revised{13.11} & \revised{47.33} \\
\texttt{phi3v-ft-nf} & 0.489(0.493/0.485) & 0.516(0.531/0.499) & 0.465(0.460/0.472) & \revised{12.67} & \revised{47.21} \\
\texttt{phi3v-ft-mq} & 0.487(0.490/0.484) & 0.511(0.527/0.494) & 0.465(0.457/0.474) & \revised{12.68} & \revised{47.03} \\
\texttt{phi3v-allft} & \uline{0.531}(\uline{0.549}/\dotuline{0.510}) & \uline{0.555}(\uline{0.583}/\uline{0.523}) & \dotuline{0.509}(\uline{0.518}/0.497) & \revised{\uline{13.69}} & \revised{\uline{48.06}} \\
\texttt{phi3v-allft-nf} & \dotuline{0.526}(\dotuline{0.538}/\uline{0.512}) & \dotuline{0.548}(\dotuline{0.574}/\dotuline{0.519}) & 0.505(\dotuline{0.506}/\dotuline{0.505}) & \revised{\dotuline{13.57}} & \revised{\dotuline{47.88}} \\
\texttt{phi3v-allft-mq} & 0.519(0.531/0.504) & 0.543(0.567/0.516) & 0.496(0.500/0.492) & \revised{13.19} & \revised{47.55} \\\bottomrule
\end{tabularx}
\end{table}

Second, Table \ref{table:result_ingproc} presents the results of the comparative evaluation of ingredients and cooking procedures between the generated recipes and the ground truth recipes.
We firstly focus on the evaluation results of the ingredients.
Looking at the micro F1 values, the results of training LLaVA-1.5 models show performance comparable to GPT-4o, and the Phi-3 Vision model, when fine-tuned with full parameters for its LLM part, even surpasses GPT-4o in performance.
The highest micro F1 score was achieved by fine-tuning Phi-3 Vision, including the image encoder, with only recipe data.
However, looking at precision and recall, GPT-4o has a higher recall than precision and achieves the highest recall value, while the models fine-tuned in this research have higher precision than recall, indicating a different tendency between the models.
Focusing on the performance difference between seasonings and other ingredients, GPT-4o can output seasonings more accurately than non-seasoning ingredients, while the fine-tuned models tend to be more accurate with non-seasoning ingredients.
When comparing between models, as the overall performance improves, the performance of non-seasoning ingredients improves more than that of seasonings, suggesting that it is easier to learn the differences in ingredients that are visually apparent in the image than the differences in seasonings, which are less visually apparent.
It is conjectured that seasonings are generated based on the knowledge of LLMs, such as the title of the recipe and the compatibility with other ingredients that have been output earlier, which may also lead to the tendency for seasonings to have higher recall than other ingredients.
Moving on to the comparison between training data, LLaVA-1.5 models shows better ingredient performance when fine-tuned with additional Non-food data, while the Phi-3 Vision model shows worse ingredient performance when fine-tuned with additional Non-food data.
These differences are likely due to the original performance, the amount and content of training data, and the size of the LLM part, of the base MLLMs.

Next, we focus on the evaluation results of the cooking procedures.
\revised{The \texttt{phi3v-allft} showed the best performance in both sacreBLEU and ROUGE-L, while GPT-4o showed the worst.}
Looking at the trained models, both LLaVA-1.5 and Phi-3 Vision have sacreBLEU scores around \revised{10} and ROUGE-L scores around \revised{45}, but the Phi-3 Vision model with full parameter fine-tuning of the LLM part shows \revised{an improvement of about three points} in both scores.
This suggests that the performance of cooking procedures is greatly influenced by the language capabilities of the MLLM.

\begin{figure}[t]
\centering
\includegraphics[width=0.9\textwidth]{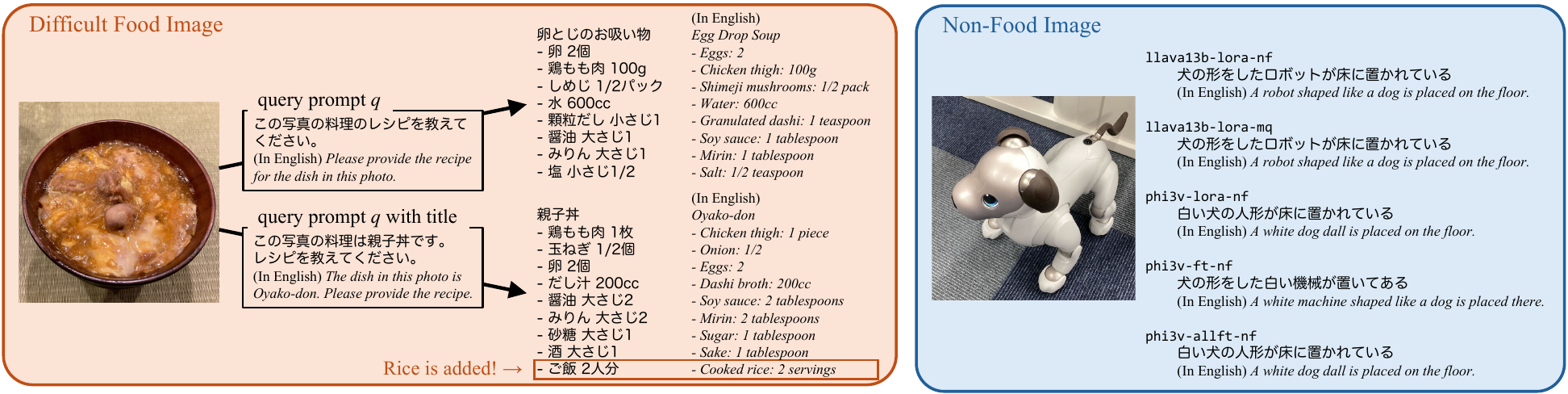}
\caption{Example outputs of our models. \textbf{\color[rgb]{0.753,0.310,0.082} Left}: the difficult food example. \textbf{\color[rgb]{0.133,0.369,0.600} Right}: the non-food image example.}\label{fig:example}
\end{figure}

Finally, we present the results of applying our model to images in Figure \ref{fig:example}, which are not publicly available on the internet and new to LMMs.
While Figure \ref{fig:overview} shows the output of \texttt{llava13b-lora-mq}, we introduce two more examples in Figure \ref{fig:example} that better illustrate the model's performance.
The left side of the figure shows an example that is difficult to judge by appearance alone and answers from \texttt{phi3v-allft-mq} model for it.
This dish contains rice, but it is difficult to distinguish from the photo.
It is shown in the top recipe in the figure that the recipe generated from the photo does not include rice as an ingredient.
However, the result is different when using the MLLM trained by R/MQ includes the dish name in the query text $q$ which is given by the user in addition to the photo.
We can see in the lower recipe in the figure that the generated recipe for the photo and its dish name includes rice.
The right side of the figure shows the difference in output between models when a non-food photo is input.
All models recognize that the photo is a dog, but there are subtle differences.

\section{Conclusion}
In this research, we have developed the Japanese recipe text generation model from food images.
We have focused on developing more practical models by incorporating both food and non-food images and experimenting with the Phi-3 Vision model, which has only around 4B parameters.
By utilizing a proposed LLM-based evaluation metric, our model has demonstrated superior ingredient generation performance compared to GPT-4o.
Furthermore, by training the model under various conditions and evaluating it from multiple perspectives, we have gained valuable insights into the understanding of MLLMs.

Future directions for this research include developing an LLM-based evaluation framework for ingredient quantities and cooking procedures, which were not analyzed in detail in this research, and investigating the feasibility of the generated recipes through actual human cooking experiments.
Additionally, we envision potential applications of our trained model, such as providing initial values for recipe registration on recipe-sharing websites, incorporating it into food logging and management systems, and utilizing it for nutrient estimation.

\begin{credits}
\subsubsection{\discintname}
The authors have no competing interests to declare that are relevant to the content of this article.

\subsubsection{\ackname} This research was partially supported by JST JPMJCR22U4, JSPS KAKENHI 23K25247 and foo.log Inc.
We used ``Rakuten Dataset'' (\url{https://rit.rakuten.com/data_release/}) provided by Rakuten Group, Inc. via IDR Dataset Service of National Institute of Informatics.
\end{credits}

\bibliographystyle{splncs04}
\bibliography{main}
\clearpage

\setcounter{page}{0}
\pagenumbering{roman}
\setcounter{table}{0}
\renewcommand{\thetable}{A\arabic{table}}
\appendix
\section{Revision of Procedure Results}
In our original report, we discovered that there was an error in the calculation method of the procedure evaluation metrics.
This mistake has been corrected in this revised version.
The error was caused by an incorrect reference dataset in the procedure evaluation: we mistakenly included all the components of the recipe, the title, the ingredients, the procedure, in the reference texts.
The correct total token length of the reference dataset is 494,000, while we evaluated with the reference data with a total of 767,400 tokens.

Table \ref{table:revised_procedure} shows the comparison of two procedure metrics between in the original report and in this revised version.
The table also shows the total token length of the 5,000 generated procedure texts.
As shown in the table, the GPT-4o model produced longer outputs, whereas our model generated more concise responses.
Consequently, the scoring error disproportionately affected our model's performance, making its scores appear lower than they should have been.
Upon correction, we confirmed that our mistake had disadvantaged our model.
Importantly, this correction not only supports our original claim that our trained model can outperform GPT-4o, but actually strengthens it.
Since the numerical error did not change the overall trend of the results, we decided not to withdraw the paper.
Instead, we are publishing this revised version with corrected values and minor adjustments.
We sincerely apologize for this error and appreciate the understanding of the research community.

\begin{table}[bph]
\caption{Summary of the procedure metrics with comparison of the scores in the original report and this revised version. The underlined number indicates the best score and the dotted underlined number indicates the second-best score.}\label{table:revised_procedure}

\renewcommand{\arraystretch}{0.96}
\begin{tabularx}{\textwidth}{@{}tDDDDD@{}} \toprule
 & \multicolumn{2}{s}{original} & \multicolumn{2}{s}{\revised{revised}} & \\\cmidrule(l){2-3}\cmidrule(l){4-5}
Model name & \tiny sacre\scriptsize BLEU & \tiny ROUGE-\scriptsize L & \revised{\tiny sacre\scriptsize BLEU} & \revised{\tiny ROUGE-\scriptsize L} & \#tokens \\\midrule
\texttt{gpt-4o} & \uline{7.223} & \uline{40.24} & \revised{8.22} & \revised{41.72} & 661,609 \\\midrule
\texttt{llava7b-lora} & 3.872 & 34.60 & \revised{8.83} & \revised{43.98} & 322,582 \\
\texttt{llava7b-lora-nf} & 4.215 & 35.00 & \revised{9.41} & \revised{44.46} & 330,476 \\
\texttt{llava7b-lora-mq} & 4.603 & 35.62 & \revised{10.04} & \revised{45.02} & 340,091 \\
\texttt{llava13b-lora} &  4.205 & 35.19 & \revised{9.37} & \revised{44.61} & 331,797 \\
\texttt{llava13b-lora-nf} & 4.579 & 35.53 & \revised{9.99} & \revised{44.86} & 339,986 \\
\texttt{llava13b-lora-mq} & 4.775 & 35.87 & \revised{10.14} & \revised{45.00} & 351,195 \\\midrule
\texttt{phi3v-lora} & 4.431 & 35.16 & \revised{10.08} & \revised{45.01} & 324,441 \\
\texttt{phi3v-lora-nf} & 4.396 & 35.05 & \revised{9.98} & \revised{44.86} & 325,193 \\
\texttt{phi3v-lora-mq} & 4.207 & 34.80 & \revised{9.63} & \revised{44.49} & 322,112 \\
\texttt{phi3v-ft} & 5.945 & 36.92 & \revised{13.11} & \revised{47.33} & 337,201 \\
\texttt{phi3v-ft-nf} & 5.633 & 36.63 & \revised{12.67} & \revised{47.21} & 329,071 \\
\texttt{phi3v-ft-mq} & 5.732 & 36.71 & \revised{12.68} & \revised{47.03} & 335,475 \\
\texttt{phi3v-allft} & \dotuline{6.261} & \dotuline{37.48} & \revised{\uline{13.69}} & \revised{\uline{48.06}} & 340,409 \\
\texttt{phi3v-allft-nf} & 6.185 & 37.38 & \revised{\dotuline{13.57}} & \revised{\dotuline{47.88}} & 339,074 \\
\texttt{phi3v-allft-mq} & 6.006 & 37.15 & \revised{13.19} & \revised{47.55} & 338,070 \\\bottomrule
\end{tabularx}
\end{table}
\end{document}